\documentclass{article}

\usepackage[preprint]{neurips_2026}

\usepackage[utf8]{inputenc}
\usepackage[T1]{fontenc}
\usepackage{hyperref}
\usepackage{url}
\usepackage{booktabs}
\usepackage{amsfonts}
\usepackage{amsmath}
\usepackage{microtype}
\usepackage{xcolor}
\usepackage{graphicx}
\usepackage{tabularx}

\newcolumntype{Y}{>{\raggedright\arraybackslash}X}

\title{Thinking Costs Tokens: When More Structure is Worth the Price}

\author{
  Thomas Nolasque\thanks{This work was completed at Royal Bank of Canada as part of the RBC Amplify program.} \\
  Royal Bank of Canada \\
  \texttt{thomasnolasque@gmail.com}
  \And
  John Grey \\
  Royal Bank of Canada \\
  \texttt{john1grey9@gmail.com}
  \And
  Calista Pham \\
  Royal Bank of Canada \\
  \texttt{calistapham13@gmail.com}
  \And
  Ankit Vani \\
  Royal Bank of Canada \\
  \texttt{ankit.vani@borealisai.com}
}

\begin{document}

\maketitle

\begin{abstract}
Adding inference structure to a language model lets it search, verify, and revise, but these actions consume the very budget they are supposed to use well.
In this paper, we investigate whether there exists a token-budget threshold, below which the overhead of planning and verification hurts performance and above which it helps.
We evaluate two systems on FinQA and TAT-QA financial reasoning tasks, using GPT-5.4 mini across 14 budget tiers ranging from 250 to 42{,}000 output-equivalent tokens.
The first system is a monolith, which is a single LLM call.
The second is a verified search architecture that adds planning, label-blind checking, and repair capabilities.
We run 1{,}000 cases for a total of 28{,}000 completed cells.
Both systems score 0\% at the two lowest tiers, where neither can fit a complete prompt.
At 1{,}000 tokens, the monolith reaches 18\% accuracy while verified search scores near 0\%, since the planning overhead leaves no room for an answer.
From 1{,}500 tokens onward, verified search surpasses the monolith and maintains a consistent advantage, reaching approximately 44\% at the highest tiers while the monolith reaches approximately 40\%.
The crossover occurs between 1{,}000 and 1{,}500 output-equivalent tokens, confirmed by a strict intersection-union test ($p \le 0.001$ at both endpoints).
\end{abstract}

\section{Introduction}

Language model systems increasingly rely on test-time mechanisms such as planning, retrieval, verification, and repair \citep{wei2022chainofthought,yao2023react,lewis2020rag}.
These mechanisms can improve answer quality, but they also consume tokens.
Every planning call repeats instructions and evidence, every candidate reserves completion space, and coordination overhead can crowd out the actual reasoning it was intended to support.
Prior work shows that additional test-time compute can help, but that its best allocation depends on task difficulty \citep{snell2024testtime}.
Self-correction is also unreliable without an informative external signal \citep{kamoi2024selfcorrection}.
These observations suggest that extra inference structure is conditionally beneficial.

We aim to determine, does extra inference structure pay for itself only above a certain token-budget threshold?
We test this on financial reasoning tasks where questions require multi-step arithmetic over tabular and textual evidence.
As independent variables, we have the inference architecture (monolith versus verified search) and the token-budget tier (14 levels from 250 to 42{,}000 output-equivalent tokens).
As dependent variable, we have correctness of system's answer.

We have two main hypotheses based on the structure of the competing systems.

Hypothesis 1a:\\The monolith achieves reasonable accuracy at a lower minimum token budget than verified search.
This is due to the monolith making exactly one retrieval call and one answer call, spending its entire budget on producing a direct response.
However, verified search must first run a planner, then retrieve, generate candidates and check, which at low budgets, leaves too few tokens for the answer itself.

Hypothesis 1b:\\Verified search outperforms the monolith once the budget is large enough to support its full mechanism.
This is because planning can improve evidence selection, multiple candidates increase the chance of a correct draft, and the label-blind checker can reject unsupported arithmetic before a wrong answer is accepted.
Thus, at high budgets, the consumed tokens being better utilized for improving accuracy, and the mechanisms fit within the allocation.

Together, these hypotheses predict a crossover, defined as a budget region below which the monolith has a better return on its token investment and above which verified search has a better return.

Let $D_b$ be the paired accuracy difference, verified search minus monolith, at budget tier $b$.
The confirmatory hypothesis is
\begin{equation}
H_1: D_{\mathrm{low}} < 0 \quad\text{and}\quad D_{\mathrm{high}} > 0.
\label{eq:hypothesis}
\end{equation}
We rely on both directional components holding, since a positive high-tier effect alone would be a threshold benefit, not a crossover.

\section{Related work}

\subsection{Test-time compute scaling}

\citet{snell2024testtime} showed that scaling test-time computation can be more effective than scaling model parameters.
However, optimal allocation depends on task difficulty, such that easy problems benefit less from additional compute, while hard problems see meaningful improvements.
This leads us to design our budget-tier design such that we vary resources across a wide range to identify where additional structure begins to improve capability, compared to simply scaling up the monolith.

\subsection{Self-correction in language models}

\citet{kamoi2024selfcorrection} surveyed self-correction in large language models and found it ineffective without external feedback signals.
LLMs that attempt to correct their own outputs without grounded verification often degrade performance instead of improving it.
Our verified search system handles this by using a label-blind checker that validates arithmetic and citation provenance, providing an informative external signal rather than relying on self-assessed confidence.

\subsection{Chain-of-thought and structured reasoning}

\citet{wei2022chainofthought} showed that prompting large language models with intermediate reasoning steps and chain-of-thought prompting, substantially improves performance on arithmetic, commonsense, and symbolic reasoning tasks.
\citet{yao2023tot} generalized this idea of Tree of Thoughts, a framework that explores multiple reasoning paths with self-evaluation and backtracking, enabling deliberate planning at test time.
\citet{wang2023selfconsistency} proposed self-consistency decoding, where we sample diverse reasoning paths and select the most consistent answer by marginalizing over them, yielding large gains on arithmetic benchmarks.
Our verified search system draws on similar principles, since we generate multiple candidate solutions and use a checker to select among them, rather than relying on a single greedy generation.

\subsection{Retrieval-augmented generation}

\citet{lewis2020rag} introduced retrieval-augmented generation (RAG), combining a parametric language model with a non-parametric dense retrieval index to ground generation in external evidence.
RAG showed that augmenting language models with retrieved passages improves factual accuracy on tasks that require knowledge.
Our systems use a deterministic retriever to supply financial evidence at test time, with the key difference being that we study how retrieval interacts with a fixed token budget, since each retrieval round consumes tokens that could otherwise be spent on answer generation.

\subsection{Agentic reasoning and tool use}

\citet{yao2023react} proposed ReAct, which interleaves reasoning traces with actions for specific tasks such as search queries, allowing language models to plan, retrieve, and revise within a single inference trajectory.
ReAct demonstrated that combining reasoning and acting reduces hallucination and error propagation, relative to reasoning-only or acting-only baselines.
Our verified search architecture shares this interleaved design, since it plans queries, retrieves evidence, generates candidates, and repairs failures, all within a single budget-constrained execution, but we focus on the cost side of this design and whether the overhead pays for itself.

\subsection{Iterative refinement and repair}

\citet{madaan2023selfrefine} introduced Self-Refine, which uses a single LLM to generate an initial output, provide feedback, and iteratively refine it without additional training.
\citet{shinn2023reflexion} proposed Reflexion, where agents verbally reflect on task feedback and store reflections in episodic memory to improve subsequent attempts.
Both approaches show that iterative revision can improve output quality, but they do not account for the cumulative token cost of each refinement cycle.
Our repair mechanism is budget-aware, due to it using checker findings as the feedback signal and only attempting repairs when the remaining token budget permits, making a crisp cost-benefit tradeoff.

\subsection{Process verification}

\citet{lightman2023letsverify} showed that process supervision, providing feedback on each intermediate reasoning step, rather than only on the final answer, significantly outperforms outcome supervision, when the training rewards models for mathematical reasoning.
This finding supports our use of a label-blind checker that validates intermediate arithmetic and citation provenance, rather than only checking the final answer.
The checker in our verified search system functions as a lightweight process verifier, operating at test time without access to gold labels.

\subsection{Financial question answering}

FinQA \citep{chen2021finqa} pairs financial-report questions with text, tables, and executable reasoning programs.
TAT-QA \citep{zhu2021tatqa} similarly combines tabular and textual financial evidence with arithmetic and counting questions.
Both datasets require multi-step numerical reasoning over structured evidence, aimed at measuring whether sequential retrieval and verification improve correctness compared to a single-call approach.

\section{Methodology}

\subsection{Datasets and selection criteria}

As datasets, we use local snapshots of FinQA and TAT-QA.
The data adapter verifies each snapshot hash, safely executes the annotated derivation, locates every operand in the public evidence, and rejects unsupported operations, ambiguous scales, missing support, duplicate documents, and answer-program disagreements.
At most one question is selected from each source document.

We have selected cases that require at least two derivation operations, since this is where multi-step reasoning is necessary and where extra inference structure could improve correctness.
We use 1{,}000 balanced cases for statistical strength, with 500 drawn from FinQA and 500 from TAT-QA, all of which are deterministically chosen and independent across documents.

Public cases contain only the question, evidence corpus, difficulty tier, document identity, and descriptive metadata.
Hidden labels contain the typed answer, executable gold derivation, intended evidence item IDs used for each derivation step, and source name, document ID along with question ID.
Numeric scoring uses decimal arithmetic with explicit unit and scale normalization, entity and period compatibility checks, along with frozen absolute and relative tolerances.
A correct output requires the candidate solution's numeric string to be an exact match.

\subsection{Systems}

Every system uses the same core instructions, evidence serialization, strict candidate schema, exact model (gpt-5.4-mini), and deterministic retriever, with the only manipulated factor being the inference structure.

\begin{table}
  \caption{System mechanisms, such that both systems share the same model and retriever, the difference being in how they allocate their token budget.}
  \label{tab:systems}
  \centering
  \begin{tabularx}{\linewidth}{@{}lY@{}}
    \toprule
    System & Mechanism \\
    \midrule
    Monolith &
    Issues the question as a direct query, retrieves up to the tier limit, making exactly one answer call capped at 256 output tokens.
    It never checks or revises. \\
    Verified search &
    Runs a mandatory planner (capped at 128 tokens), performs deterministic query union over planned queries, generates sequential candidates (each capped at 256 tokens), applies label-blind checking, and accepts the first passing candidate.
    At higher tiers, it may attempt repairs using checker findings only, with the repair limit scaling alongside the token cap (see Table~\ref{tab:budgets}). \\
    \bottomrule
  \end{tabularx}
\end{table}

The checker verifies citation existence, numerical data source, safe arithmetic, candidate-expression agreement, units, scale, entity, period, and division safety.
It cannot see correctness or a gold value.
A cell is one execution of a given system on a given case at a given budget tier.
Every cell records planned and actual queries, retrieval IDs before and after truncation, candidates, checks, repair attempts, accepted index, authoritative usage, and an exit reason.

Invalid output, architecture-caused tool failure and budget exhaustion all score as incorrect.

\subsection{Budget intervention}

The hard budget is the sum of the prompt tokens and completion tokens over all calls in a cell:
\begin{equation}
T_{csb} = \sum_i (P_{csbi} + C_{csbi}) \le B_b.
\label{eq:budget}
\end{equation}
Here $T_{csb}$ is the total token usage for case $c$, system $s$, at budget tier $b$; $P_{csbi}$ and $C_{csbi}$ are the prompt and completion tokens for call $i$ within that cell; and $B_b$ is the token cap for tier $b$.

Raw token counts do not reflect computational cost, since input and output tokens are priced differently.
We therefore express budgets in output-equivalent tokens by weighting each input token by the input-to-output price ratio.
For GPT-5.4 Mini (\$0.75 per million input tokens, \$4.50 per million output tokens), this ratio is $0.17$, so one input token counts as $0.17$ output-equivalent tokens.
This makes the budget correlate more closely with actual price and resource usage than a raw token count would.

Before each call, the ledger obtains an exact token count of the prompt, reserving the maximum output, refusing any call that would exceed the remaining budget.
After the response, the ledger releases unused reservation and records the authoritative usage.
Prompt mismatch, output-reservation overrun, and hard-cap overrun are treated as protocol violations.

\begin{table}
  \caption{Budget tiers. Action limits scale with the token cap. The lowest tiers are for establishing a floor effect by being too small for a complete answer.}
  \label{tab:budgets}
  \centering
  \begin{tabular}{lrrrrr}
    \toprule
    Tier & Token cap & Retrieval & Queries & Candidates & Repairs \\
    \midrule
    t250   &    250 &  1 & 1 & 1 & 0 \\
    t500   &    500 &  1 & 1 & 1 & 0 \\
    t1000  &  1{,}000 &  2 & 1 & 1 & 0 \\
    t1500  &  1{,}500 &  2 & 1 & 1 & 0 \\
    t2000  &  2{,}000 &  3 & 1 & 1 & 0 \\
    t3000  &  3{,}000 &  3 & 2 & 2 & 1 \\
    t5000  &  5{,}000 &  4 & 2 & 2 & 1 \\
    t8000  &  8{,}000 &  6 & 3 & 3 & 2 \\
    t12000 & 12{,}000 &  8 & 4 & 4 & 2 \\
    t18000 & 18{,}000 & 10 & 5 & 5 & 3 \\
    t24000 & 24{,}000 & 12 & 6 & 6 & 4 \\
    t30000 & 30{,}000 & 14 & 7 & 7 & 4 \\
    t36000 & 36{,}000 & 16 & 8 & 8 & 5 \\
    t42000 & 42{,}000 & 18 & 9 & 9 & 5 \\
    \bottomrule
  \end{tabular}
\end{table}

\subsection{Confirmatory inference}

Confirmation uses an intersection-union test.
Both directional components of the hypothesis in Equation~\ref{eq:hypothesis} must pass their one-sided exact McNemar tests at $\alpha = .05$.
A crossover is confirmed only if the low difference is strictly negative, the high difference is strictly positive, and both tests reject.
The analysis is designed with a 90\% probability (statistical power) of detecting a 5\% difference.
We calculated the required sample size using early, blinded data from the initial phase of the study.

\section{Results}

\subsection{Execution summary}

We ran 1{,}000 main cases across 2 systems with 14 budget tiers per cell, yielding 28{,}000 scheduled cells.
All calls used the resolved model \texttt{gpt-5.4-mini-2026-03-17-eastus-dz}, with usage reporting through an LLM gateway API endpoint.

\subsection{Accuracy by system and tier}

\begin{figure}
\centering
\includegraphics[width=\linewidth]{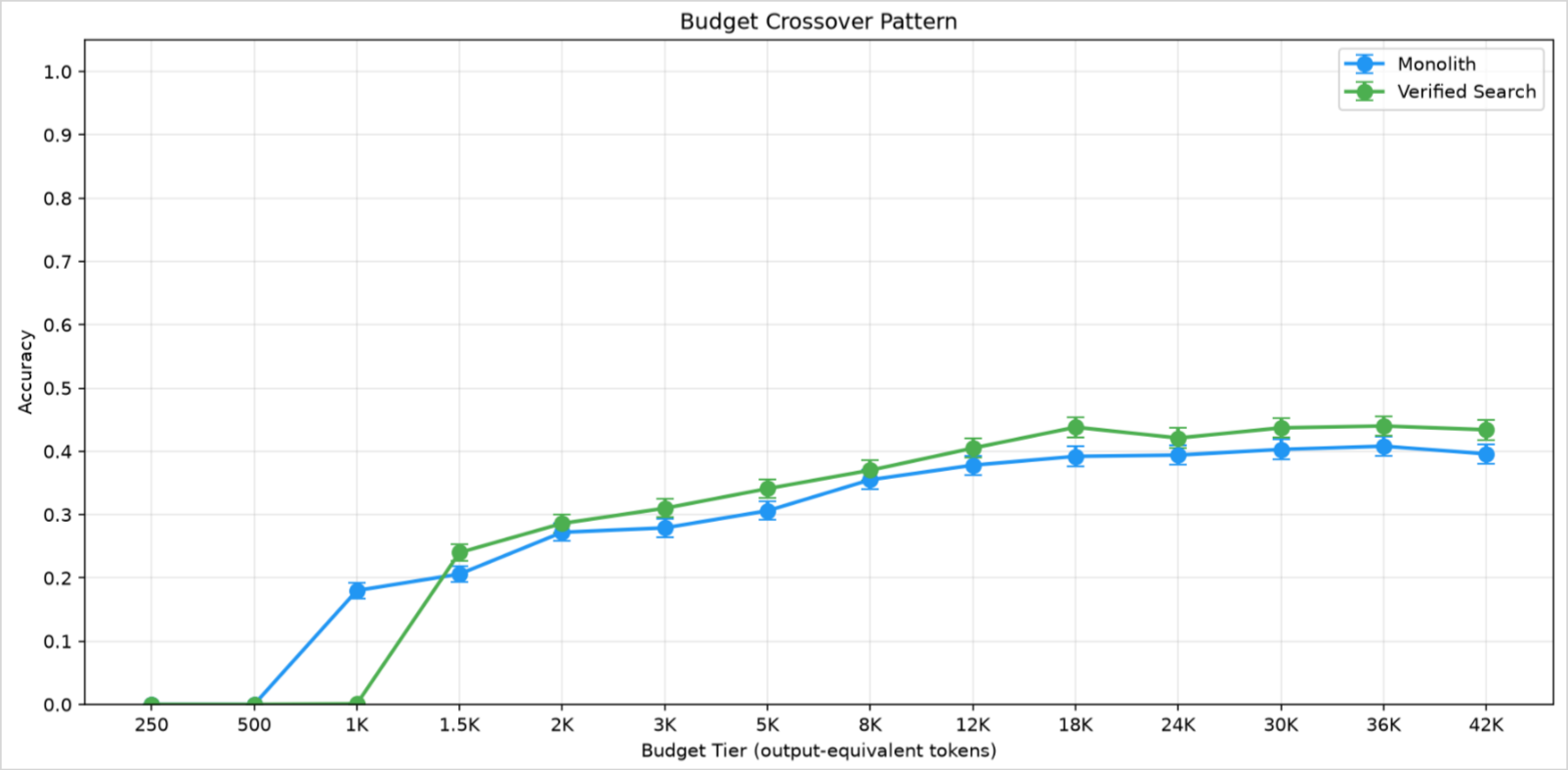}
\caption{Accuracy by system and budget tier.
At 1{,}000 tokens the monolith (blue) leads at 18\% while verified search scores near 0\%.
From 1{,}500 tokens onward verified search (green) overtakes the monolith and both systems improve continuously, reaching approximately 40\% and 44\% respectively at t42000.
Error bars show standard error across 1{,}000 cases.}
\label{fig:crossover}
\end{figure}

Both systems score 0\% at the t250 and t500 tiers, since at these budgets, neither system can fit a complete prompt and produce a valid answer before exhausting the token cap.
This confirms that the lowest tiers impose a genuine resource constraint.

At t1000, the monolith reaches 18\% accuracy (180/1000 correct) while verified search scores only 0.1\% (1/1000 correct).
At this tier, the monolith can fit exactly one retrieval call and one answer call, whereas verified search exhausts its budget on the mandatory planning call, leaving insufficient tokens to generate a candidate.

At t1500, verified search overtakes the monolith for the first time, achieving 24\% versus 20.6\%.
This is the crossover point, the minimum budget at which the planning-retrieval-candidate pipeline fits within the allocation.
From t1500 onward, verified search maintains a consistent advantage over the monolith at every tier.

Both systems continue improving as the budget grows.
The monolith rises from 18\% at t1000 to approximately 40\% at t42000, benefiting from additional retrieval context even with a single answer call.
Verified search rises from 24\% at t1500 to approximately 44\% at t42000.
At higher tiers, multiple retrieval rounds, multiple candidates, and repair cycles all become affordable, with the checker rejecting incorrect first drafts and the repair mechanism attempting corrections.

\subsection{The crossover region}

At t1000, the monolith achieves 18\% accuracy while verified search scores near 0\%.
At t1500, verified search reaches 24\% while the monolith reaches 20.6\%.
The crossover thus occurs between t1000 and t1500 output-equivalent tokens.

This is consistent with both hypotheses.
The monolith achieves reasonable accuracy at a lower minimum budget because it spends everything on a single retrieval and answer call.
Verified search requires at least 1{,}500 tokens before its planning-retrieval-candidate pipeline fits within the allocation, but once the budget supports this pipeline, it converts the additional structure into higher accuracy at every subsequent tier.

The confirmatory intersection-union test confirms the crossover.
At the low endpoint (t1000), the paired McNemar test rejects with $p \le 0.001$ (180 discordant pairs favoring monolith, 1 favoring verified search).
At the high endpoint (t1500), the test rejects with $p \le 0.001$ (78 pairs favoring verified search, 44 favoring monolith).
Both directional components hold, confirming a strict crossover per Equation~\ref{eq:hypothesis}.

\subsection{Mechanism traces}

The cell-level traces confirm that the accuracy pattern reflects genuine mechanism differences rather than random variation.
At t1000, verified search averages 1.01 calls and produces only 0.01 candidates per cell, meaning it exhausts its budget on the planning call alone, leaving insufficient tokens for a candidate.
At t1500, it averages 2.0 calls and 1.0 candidate, indicating the pipeline first fits at this tier.
At t8000 and above, verified search averages 2.6 calls and 1.4 candidates, with 12-19\% of cells using the repair mechanism.
The monolith consistently makes exactly 1 call and 1 candidate at t1000 and above, confirming that additional budget beyond its minimum provides no structural advantage, only richer retrieval context.

\subsection{Floor effects}

At t250 and t500, both systems score 0\%.
At these tiers, neither system can fit a complete prompt and answer within the budget, confirming a genuine resource floor.
The crossover occurs immediately above this floor, such that at t1000 the monolith activates while verified search cannot yet fit its pipeline, and by t1500 both systems are active with verified search already ahead.

\section{Discussion}

The results confirm the conditional crossover hypothesis.
At low token budgets (t1000), the monolith is the better investment, achieving 18\% accuracy while verified search does not produce a candidate.
At higher budgets (t1500 onward), verified search is the better investment, converting the additional tokens into planning, broader retrieval, and verification that the monolith cannot leverage.

The crossover between t1000 and t1500 corresponds precisely to the budget at which verified search's planning-retrieval-candidate pipeline first fits within the allocation.
Below t1500, the architecture has the scaffolding but not the budget to execute it.
Above t1500, the scaffolding pays for itself immediately and consistently.

Contrary to our initial expectation, the monolith does not plateau.
It improves continuously from 18\% at t1000 to approximately 40\% at t42000, benefiting from richer retrieval context at higher tiers even with a single answer call.
However, verified search improves faster, reaching approximately 44\% at t42000.
The gap between systems remains positive (3-5 percentage points) at every tier from t1500 onward, indicating that the structural advantage of planning and verification is not a one-time benefit but compounds with available budget.

For applications where the token budget is constrained below approximately 1{,}500 output-equivalent tokens, a single-call monolith is the more efficient architecture.
For applications where the budget can reach 1{,}500 tokens or more, verified search is optimal, since it uses the additional budget productively from its very first feasible tier onward.

\subsection{Limitations}

All systems use the same underlying model (GPT-5.4 Mini), so correlated errors across systems are expected.
A different model could shift the crossover point or the magnitude of the effect.
All calls use the default temperature (1.0) with a single sample per call, thus, majority-vote decoding over multiple samples could interact with budget allocation in ways this study does not measure.
Similarly, results may be sensitive to the specific prompt wording, since no ablation over prompt variants was conducted.

We do not explore hybrid or adaptive systems that switch strategy based on available budget, since we only compare two fixed architectures.
The mapping from budget tier to action limits (retrieval rounds, candidates, repairs) in Table~\ref{tab:budgets} was hand-designed, meaning a learned or adaptive allocation policy might perform better.
The label-blind checker implements a specific set of validation rules (citation existence, arithmetic safety, unit compatibility), thus a different checker design could shift verified search's accuracy curve.

Both systems use the same deterministic BM25 retriever.
A learned dense retriever could raise the retrieval quality ceiling for both systems, potentially shifting the crossover point or narrowing the gap between architectures.

The hard budget measures gateway-reported prompt and completion tokens, but tool calls, CPU time, wall time, latency, and monetary cost were logged, but not analyzed, despite being potentially relevant.
Scoring uses exact numeric match, providing no partial credit, such that a softer metric might reveal different tradeoffs.
We report aggregate accuracy without stratifying by case difficulty, meaning the crossover might occur at different budgets for easy versus hard cases.
No human expert baseline is reported, making it difficult to contextualize the absolute accuracy levels (18-44\%) against an upper bound.

The study is limited to financial reasoning tasks from FinQA and TAT-QA.
Whether the crossover generalizes to other domains with different evidence structures, question types, or reasoning depths remains an open question.
Future work could expand both the domain coverage and the token budget range to sharpen the transitions from floor effects to meaningful accuracy.

\section{Conclusion}

We investigated whether extra inference structure in language model architectures pays for itself only above a token-budget threshold, and we concluded that it does, conditionally.
Below approximately 1{,}500 output-equivalent tokens, the monolith provides a better return on its token budget, since verified search cannot fit its planning-retrieval-candidate pipeline within the allocation.
At 1{,}500 tokens and above, verified search provides a better return, converting additional tokens into broader evidence coverage, multiple candidate attempts, and label-blind verification that catches errors before the answers are submitted.

The crossover at 1{,}000-1{,}500 tokens is confirmed by a strict intersection-union test ($p \le 0.001$ at both endpoints) over 1{,}000 cases.
Architecture selection should be conditional on the available token budget, since a system that helps at 1{,}500 tokens may be completely inoperative at 1{,}000.

\begin{ack}
We thank the developers of FinQA and TAT-QA for making their datasets publicly available.
This work was completed as part of the RBC Amplify program at Royal Bank of Canada.
\end{ack}

\bibliographystyle{plainnat}
\bibliography{references}

\end{document}